# Learning Mobile App Usage Routine through Learning Automata

Ramin Rahnamoun[1], Reza Rawassizadeh[2], Arash Maskooki[3]
[1]Computer Engineering Department, Islamic Azad University, Central Tehran Branch, Tehran, Iran
[2]Computer Science Department, Dartmouth College, NH, US
[3]Department of Electrical and Computer Engineering, UC Riverside, CA, US
r.rahnamoun@iauctb.ac.ir, rrawassizadeh@acm.org, arash.maskooki@ucr.edu

## ABSTRACT
Since its conception, smart app market has grown exponentially. Success in the app market depends on many factors among which the quality of the app is a significant contributor, such as energy use. Nevertheless, smartphones, as a subset of mobile computing devices. inherit the limited power resource constraint. Therefore, there is a challenge of maintaining the resource while increasing the target app quality. This paper introduces Learning Automata (LA) as an online learning method to learn and predict the *app usage routines* of the users. Such prediction can leverage the app cache functionality of the operating system and thus (i) decreases app launch time and (ii) preserve battery. Our algorithm, which is an online learning approach, temporally updates and improves the internal states of itself. In particular, it learns the transition probabilities between app launching. Each App launching instance updates the transition probabilities related to that App, and this will result in improving the prediction. We benefit from a real-world lifelogging dataset and our experimental results show considerable success with respect to the two baseline methods that are used currently for smartphone app prediction approaches.

### 1. INTRODUCTION & BACKGROUND
Since its conception, smart app market has grown exponentially. According to a report by statista website from 2014 [7], the number of Applications that are available are 1.3 million on Google's Market and 1.2 million for Apple's iTunes Store. There are several research works [5, 6] that show the average number of installed application (App) are more than 40 per smartphone. Similar to other mobile computing devices, smartphones suffer from limited battery power. This necessitates optimization mechanisms significant consumption contributors. One major contributor is the application load time. The launch of an App may take a few seconds [13], while the screen is on. For some resource intensive apps such as Games the application load time is even longer than average.

One solution to reduce the app search and load time is caching apps into memory. The challenge is the large memory cost of this practice. Therefore, a learning mechanism can create a favorite app list for pre-loading into cache [12].

App usage prediction is not a new problem and different algorithms [12] or frameworks [2, 11] are proposed to predict app usage. Some of these research works focused on the problem of dependencies between Apps launching sequence [5,13] or more specifically on transition probabilities between App usages [3]. Moreover, personalized App recommendation is another topic of research [1]. There are works that focused on the use of clustering algorithms [4, 5] to classify App usage based on contextual information. Context data can be included in sensors based activities such as running, walking or other human centric activities such as SMS, call, etc. [13, 6]. Some researches use time series model to tackle this problem [8].

Learning Automata (LA) is an old reinforcement learning method that is being used in a wide range of applications. Here we use Finite Action-set Learning Automata (FALA) [10], but there are several different models of LA are proposed in literature [9, 10].

All these works operate based on leveraging the temporal history of app usage as input [2, 4, 8, 11]. Some approaches [4] concurrently gather other sensor information from smartphone and correlate with app launch. By analyzing the temporal history of app usage log, these algorithms create a model from user behavior over the time and use it to predict future app launching mechanism. To the best of our knowledge, most of the existing methods [1, 4, 15, 18] have the offline approach and do not perform the learning process on the device. In other words, these methods solve an *offline* classification problem and they classify this data in a cloud, and not on the smartphone. This characteristic imposes the burden of network reliance, privacy and response time associated with transferring the data and receiving the result. As it has been described due to size limitations of smartphones, they cannot execute computationally complex algorithms.

Some other algorithms [15,16,17], use different sensors as data sources. Using more than one sensor for prediction provides a better accuracy in comparison to focus on App usage only. However, it will impose a battery overhead, unless the algorithm is custom designed for resource efficiency [17].

In general, it is not optimal to run complex classification algorithms on mobile computing devices. We use learning automata as it is resource efficient enough [9, 10] and can operate on-device. When the number of actions are not too large, i.e. the case in app usage learning, the computational complexity is low and near linear with respect to sample size. Therefore, its response time and throughput will be acceptable to run on smartphones. Moreover, our algorithm is continuously updating itself based on analyzing temporal traces of app launching history. Therefore, there will be no challenge of concept drift [11] associated with our algorithm.

## 2. LEARNING AUTOMATA & DEFINITIONS

Learning automaton is an adaptive decision making model belong to the group of reinforcement learning methods inspired from biological system. All models of learning automata, interact with surrounding environment in discrete time instants. At each time instant, automaton randomly chooses an action, this action is a sort of input for the environment and therefore, the automaton gets a response from the environment. This reaction of environment is called *reinforcement*. The environmental response is an input to the automaton and update their states based on reward or penalty values. Formally, Learning Automata (LA) is defined by a quadruple (O, R, Q, F), where O is a set of outputs or actions that is chosen by automaton in time instant, R is a set of reinforcement values that may be discrete or continuous, Q is the set of internal state of automaton, and F: $Q \times O \times R \rightarrow Q$ is the learning algorithm and also a mapping that updates internal state of automaton.

The environment can be formally defined by triple (O, P, R), where O is the set of actions, R is the output of environment and also be the input for automaton, and P is a set of penalty probabilities. Each $p_i \in P$ relates to an action $o_i \in O$.

Learning automata can be categorized into two groups: fixed structure and variable structure [9]. The above definition relates to variable structure learning automata which is the most used structure. This paper uses Finite Action-set Learning Automata (FALA) which is a type of variable structure type [10].

FALA is defined as quadruple. Let $q_i(t)$ be the action probability in which action $o_i$ is chosen at instant t and q(t) is the vector of action probability. Suppose |O| = r, so q(t) is a vector with r elements : $q(t) = [q_1(t),\ldots,q_r(t)]$. The general form of learning algorithm is as follows: q(t+1) = F(q(t), o(t), r(t)). Here F is the learning function, o(t) is the action selected at instant t, and r(t) is the response of environment to the selected action.

The set of possible response of environment (R), may be discrete or continuous. If R = {0, 1}, it is called P-model and this paper uses this type of R set. In P-model, 0 assigns to penalty or unfavorable response and 1 assigns a reward (favorable) response.

Learning automata tries to learn optimal action from set O with respect to responses that receive from environment (set R). LA can update action probability distribution according to different formulas. We use Linear Reward-Inaction ($L_{R-I}$) algorithm [10]. Let $q(k) = q_i$, then the $L_{R-I}$ algorithm use the following update rule:

$$\begin{cases} q_i(t+1) = q_i(t) + \lambda \times r(t)(1 - q_i(t)) \\ q_j(t+1) = q_j(t) + \lambda \times r(t) q_j(t) \quad \forall j \neq i \end{cases} \quad (1)$$

In above equations, $\lambda$ is the learning parameter ($0 < \lambda < 1$), and $r(t) \in \{0, 1\}$ (use P-model environment).

Assume A = <($a_1$, $t_1$), ($a_2$,$t_2$), … > is an infinite sequence of pairs where $a_i$ represent launching of a specific App at time instant $t_j$. Therefore, $a_i \in APP \wedge t_i \in T$ , where APP is the set of installed Apps in specific user smartphone and T is the time vector. Hence, sequence A is the temporal behavior of a specific user about launching Apps.

Let $A_{ik} = (a_i, t_k) \wedge A_{ik} \in A$ and $\alpha_{ij} = P(A_{i,k+1} | A_{j,k})$ . $\alpha_{ij}$ be the conditional probability of launching App $a_i$ in time $t_{k+1}$ with the condition of launching App j in time $t_k$. The difference between k+1 and k is less than a time interval threshold ($\Delta$). $\alpha_{ij}$ is called *the transition probability* between Apps i and j.

> **Definition 1**: *Application Transition Probability Matrix (ATPM) is a matrix of* $\alpha_{ij}$. *This matrix shows the conditional probability between every pair of Apps that belong to APP.*

> **Problem:** *Given a sequence of launching event over time (A), the objective of this work is to find an estimator* $\widehat{\alpha_{ij}}$ *for* $\alpha_{ij}$.

Therefore, the proposed algorithm finds an estimator for ATPM and this process must be done based on sequence A. This sequence is not offline accessible. In other words, every App launching generates an event in smartphone and this event activate online learning algorithm to update ATPM matrix.

## 3. PROPOSED ALGORITHM

The goal of the proposed algorithm is to update $\alpha_{ij}$ values in ATPM and concurrently offer k-top Apps. Let $\alpha_i$ be the row i of ATPM. Each column of this row, shows an estimation of transition probability between App i with every other installed Apps in column j ($\alpha_{ij}$). This research uses a FALA for each ATPM rows. So we need a vector of FALAs to calculate ATPM. Let us consider the vector of FALAs as F = [$F_1$, $F_2$, … , $F_n$] and n = |APP|. Each $F_i$ has n actions, therefore $q(t) = [q_1(t),\ldots,q_n(t)]$. O is defined as launching App i after App j in a specified time interval ($\Delta$) and $o_i$ of $F_j$ shows this relation. According to equation (1), $L_{R-I}$ is used as updating rule of action probability of each automaton. If FALA offer correct App, the R value will be 1, otherwise R gets a 0 value.

FALA works with the environment and hence action probabilities update over time. Let *PrvLaunchedApp* be the last App that is launched in smartphone of a user. App launcher monitoring routine (*LaunchMonitor*) is a basic module that monitors every launching events on smartphone and triggers an event for the algorithm 1 to get the name of App. Let *NextLaunchedApp* be the current App that *LaunchMonitor* assigns new launched App to this variable.

Each cycle of the algorithm is activated when a new App launching event is monitored. In this cycle, if the time difference of *PrvLaunchedApp* and *NextLaunchedApp* is less than a threshold (Δ), *NextLaunchedApp* value is compared with the k-top Apps that offer by F with index of *PrvLaunchedApp*. $L_{R-I}$ learning algorithm update q vector based on the result of this comparison. The main loop of the algorithm is a forever loop, because the updating process of ATPM continues by each new App launching. *Algorithm 1* is the pseudo-code of this algorithm. kTOP is the ordered list of k-top Apps that is offered to the user.

**Algorithm 1. FALA for App usage routine**

**Input**: *A* (sequence of launched Apps over time)
**Output**: *ATPM, kTOP*
$\forall F_i \;:\; \forall j \; q_j(0) \leftarrow \frac{1}{|APP|}$
$PrvLaunchedApp \leftarrow LaunchMonitor()$
**for** t = $t_1, t_2, \ldots, t_k, \ldots$
  $kTOP \leftarrow k\ F_{PrvLaunchedApp}$ *with higher probabilities*
  NextLaunchedApp ← $LaunchMonitor()$
  **if** $t_{k-1} - t_k < \Delta$
    **if** NextLaunchedApp ∈ kTOP
      $F_i :\; q_i(t+1) = q_i(t) + \lambda r(t)(1 - q_i(t))$
    **else**
      $F_i :\; q_j(t+1) = q_j(t) + \lambda r(t) q_j(t)$
    **end if**
  **end if**
  PrvLaunchedApp ← NextLaunchedApp
**end for**

## 4. EXPERIMENTAL RESULTS

Dataset of this research was gathered from 35 participants, with age rage 19-22 where 18 of them were females. Further details on data collection setting is given in [19]. Each participants installs a UbiqLog [20] and runs data for 60 days period[1]. App launching frequencies of participants were different.
The implementation of the algorithm has been conducted via Matlab/Octave. The target system of running the experiment includes an Intel Core i-7 CPU with 8GB of memory.

### 4.1 Accuracy Analysis

This paper uses three performance metrics: 1) Recall@N 2) Discounted Cumulative Gain (DCG) and 3) mean Reciprocal Rank (MRR) [4] [5]:

$$Recall_K = \begin{cases} 1 \; if \text{ NextLaunchedApp} \in \text{kTOP} \\ 0 \; if \text{ NextLaunchedApp} \notin \text{kTOP} \end{cases}$$

Recall@N evaluates number of times the next launched App belong to the set of k-top ranked Apps that is offered by FALA.

[1] The dataset is available at: https://goo.gl/rXxfnu To get the code for cleaning the data please contact authors.

$$DCG = \begin{cases} 1 \; if \text{ NextLaunchedApp} = \max(\text{kTOP}) \\ \dfrac{1}{\log_2 Pos(kTOP)} \quad otherwise \end{cases}$$

In this equation Pos is a function that determines the position of current App in the list of kTOP.

$$MRR = \frac{1}{Pos(kTOP)}$$

According to Kim and Mielikäinen [4]:

> "*Reciprocal rank is another measure that discounts relevance based on the hit position. Its average, called mean reciprocal rank (MRR), is often used to assess the quality of ordered items.*"

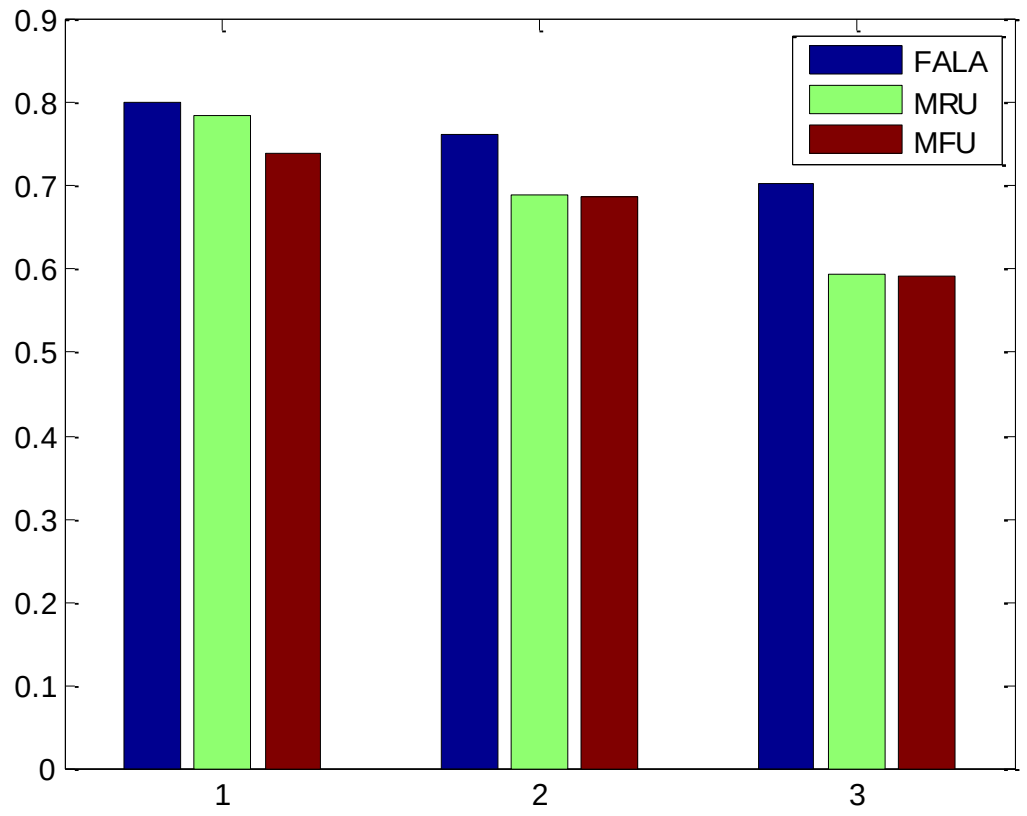

**Figure 1. Precision (vertical axis in percent) of proposed algorithm (FALA) with two other basic methods (horizontal axis). 1: Recall@6, 2: DCG, 3: MRR**

### 4.2 Results

The performance of the proposed algorithm in this paper, named FALA, is compared with two basic recommender methods that is used in different operating systems [5]:

- **Most Recently Used (MRU)** that recommend recently used Apps from most recently to least recently ones.
- **Most Frequently Used (MFU)** that evaluates frequencies of Apps and recommend them in decreasing order.

Figure 1 presents average prediction accuracy of the proposed method (FALA) and the two basic methods (MRU and MFU). Results show better App prediction accuracy of the proposed algorithm. The difference between FALA and two other methods are greater in DCG (number 2 in figure) and MRR (number 3 in figure) with respect to Recall@6. These two performance metrics relate to the position of matched result between k-top App ordered list (kTOP) and this shows that our proposed algorithm matches with top ranked Apps in the list of suggested Apps.
Figure 2 shows the App launching accuracy prediction of an anonymous user (number 13) in 60 days interval. In this figure, the proposed algorithm (FALA) can get a higher precision rate, after initial launching. This figure insists on better performance of FALA with progression of time.

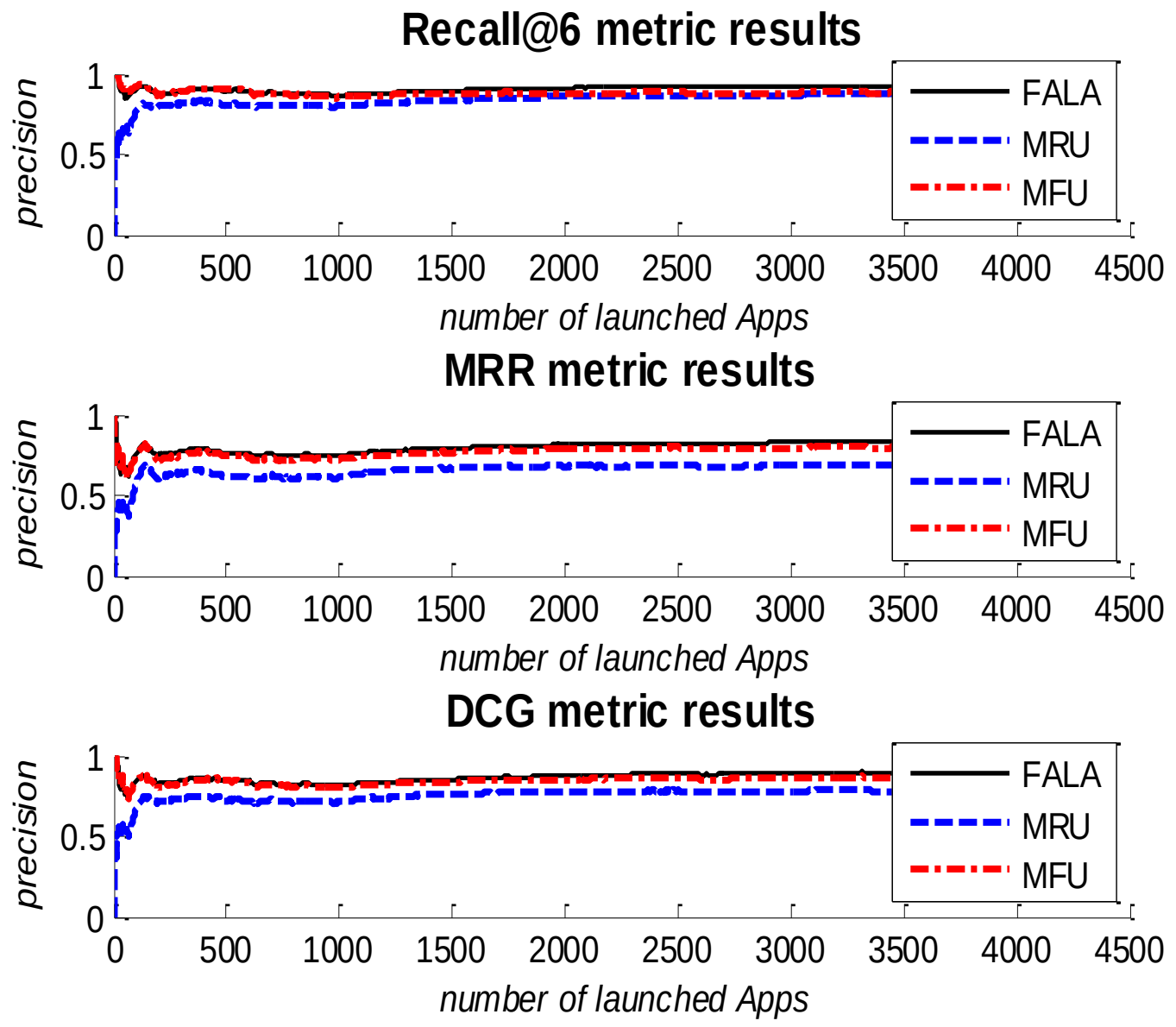

**Figure 2. Precision of three methods (y axis) over number of launched Apps (x axis). This sample belongs to user "13".**

## 5. CONCLUSION AND OUTLOOKS

This paper suggests a novel LA based algorithm to estimate transition probabilities between Apps that installed on user smartphone. The algorithm uses FALA model of LA and the experimental results show acceptable performance with respect to the basic methods (MRU and MFU). The proposed algorithm gets the launching information from a monitoring module and update its internal state for each App launching event. So, this is an online learning mechanism and does not need offline clustering process or Internet connection for data transmission. This is important because it results in less energy consumption in resource constraint devices such as smartphones.

The current research can be extended to transfer the knowledge of each FALA to other users. This leads to development of a network of FALA that can share the knowledge between them. Moreover, different models of LA (for example GLA [10]) can be tested and compared with FALA. In addition, the algorithm can be improved with other contextual information. however, this leads to extra battery usage that is an important problem in this field.